# DESIGN OF
# AN ELECTRO-HYDRAULIC SYSTEM
# USING NEURO-FUZZY TECHNIQUES

**P.J. Costa Branco** and **J.A. Dente**

Mechatronics Laboratory
Department of Electrical and Computer Engineering
Instituto Superior Técnico, Lisbon
Portugal

Increasing demands in performance and quality make drive systems fundamental parts in the progressive automation of industrial processes. Their conventional models become inappropriate and have limited scope if one requires a precise and fast performance. So, it is important to incorporate learning capabilities into drive systems in such a way that they improve their accuracy in realtime, becoming more autonomous agents with some "degree of intelligence."

To investigate this challenge, this chapter presents the development of a learning control system that uses neuro-fuzzy techniques in the design of a tracking controller to an experimental electro-hydraulic actuator. We begin the chapter by presenting the neuro-fuzzy modeling process of the actuator. This part surveys the learning algorithm, describes the laboratorial system, and presents the modeling steps as the choice of actuator representative variables, the acquisition of training and testing data sets, and the acquisition of the neuro-fuzzy inverse-model of the actuator.

In the second part of the chapter, we use the extracted neuro-fuzzy model and its learning capabilities to design the actuator position controller based on the feedback-error-learning technique. Through a set of experimental results, we show the generalization properties of the controller, its learning capability in actualizing in realtime the initial neuro-fuzzy inverse-model, and its compensation action improving the electro-hydraulics' tracking performance.

# 1 Introduction

Recent integration of new technologies involving new materials, power electronics, microelectronics, and information sciences made relevant new demands in performance and optimization procedures for drive systems. To handle command and control problems, the dynamic behavior of a drive must be modeled taking into account the electromagnetic and mechanical phenomena. However, if one requires a precise and fast performance, the control laws become more complex and nonlinear and the classical models become inappropriate and of limited scope.

The existing models are not sufficiently accurate, the parameters are poorly known, and, also, because physical effects like thermal behaviour, magnetic saturation, friction, viscosity, are in general time-variants, they are difficult to develop with the necessary simplicity and accuracy. So, it is important to develop drive systems that incorporate learning capabilities in a way that their control systems automatically improve accuracy in realtime and become more autonomous.

To investigate the possibilities of incorporating learning capabilities into drive systems, we present the implementation of a control system that uses neuro-fuzzy modeling and learning procedures to design a tracking controller to an electro-hydraulic actuator. The learning capability of the neuro-fuzzy models is employed to permit the controller to achieve actuator inverse dynamics and thus compensate the possible unstructured uncertainties to improve performance in trajectory following.

In the first part of this chapter, we present the actuator modeling using the neuro-fuzzy methodology. In this way, the information about its dynamic behaviour is expressed in a multimodel structure by a rule set composing the neuro-fuzzy model. Each region of actuator's operating domain is characterized by a rule subset describing its local behavior. The neuro-fuzzy model permits the actuator's information, codified into it, can be generalized, and use its neural-based-learning capabilities in a manner to permit modifying and/or adding knowledge to the model when necessary.

Today, conventional fuzzy controllers are publicized by industry as being "intelligent." Although, to define some "intelligence" degree, it is essential to have learning mechanisms that they do not have. Initially, some approaches have been proposed to improve the performance of conventional controllers using fuzzy logic. The first used fuzzy logic to tune gain parameters of PID controllers [34], [35], or substitute PID controllers by their fuzzy approximation [23], [36].

Some papers in the literature address control systems using learning mechanisms based on neural networks [7], [8], [12], and others introduced the idea of fuzzy learning controllers using a self-organizing approach [38], [39], or, more recently, by neuro-fuzzy structures [16], [17], [20], [37].

The second part of the chapter presents the implementation of the learning control system to the electro-hydraulic actuator combining its neuro-fuzzy inverse-model with a conventional proportional controller. This scheme results in the indirect compensation control scheme named feedback-error-learning proposed by Kawato in [5], [15], and initially explored by the authors in an unsupervized way in [18]. The controller was implemented on a Personal Computer (PC) with a 80386 CPU and an interface with A/D (analog to digital) and D/A (digital to analog) converters. All programming was done in C language, including the neuro-fuzzy algorithm and actuator signal's acquisition and conditioning.

The implemented system is constituted by real-time learning and control cycles. During these cycles, the inverse-model of the actuator uses its neural-based-learning capabilities to extract rules not incorporated into the initial model, and even change itself to characterize a possible new actuator's dynamic.

We show experimental results concerning the position control of the electro-hydraulic actuator. At each control cycle, the incorporated learning mechanism extracts its inverse-model and generates a compensation signal to the actuator. The results show that the controller is capable of generalizing its acquired knowledge for new trajectories; it can acquire and introduce new system's information in realtime using the sensor signals; and it can compensate possible nonlinearities in the system to progressively reduce its trajectory errors.

The structure of the chapter is as follows. In Section 2, the fuzzy system employed is characterized by its fuzzy logic operations. In Section 3, we review fuzzy modeling processes in the literature. Section 4 describes the neuro-fuzzy modeling algorithm. Section 5 presents the experimental system and the technique used to obtain a good training data set from the electro-hydraulic system. In Section 6, we extract the inverse-model of the actuator using the modeling algorithm presented in Section 4 and the training set of Section 5. Section 7 describes the neuro-fuzzy control system using the feedback-error-learning algorithms and presents some experimental tests.

## 2   The Fuzzy Logic System

Fuzzy sets establish a mechanism for representing linguistic concepts like big, little, small and, thus, they provide new directions in the application of pattern recognition based on fuzzy logic to automaticaly model drive systems [31], [32]. These computational models are able to recognize, represent, manipulate, interpret, and use fuzzy uncertainties through a fuzzy system.

A fuzzy logic system consists of three main blocks: fuzzification, inference mechanism, and defuzzification. The following subsections briefly explain each block, and characterize them with regard to the type of fuzzy system we used.

## 2.1 Fuzzification

Fuzzification is a mapping from the observed numerical input space to the fuzzy sets defined in the corresponding universes of discourse. The *fuzzifier* maps a numerical value denoted by $x' = (x'_1, x'_2, \ldots, x'_m)$ into fuzzy sets represented by membership functions in **U**. These functions are Gaussian, denoted by $\mu_{A^i_j}(x'_j)$ as we expressed in Equation (1).

$$\mu_{A^i_j}(x'_j) = a^i_j \exp\left[-\frac{1}{2}\left(\frac{x'_j - b^i_j}{c^i_j}\right)^2\right] \quad (1)$$

In this equation, $1 \leq j \leq m$ refers to the variable ($j$) from $m$ considered input variables; $1 \leq i \leq n_j$ considers the $i$ membership function among all $n_j$ membership functions considered for variable ($j$); $a^i_j$ defines the maximum of each gaussian function, here $a^i_j = 1.0$; $b^i_j$ is the center of the gaussian function; and $c^i_j$ defines its shape width.

## 2.2 Inference Mechanism

Inference mechanism is the fuzzy logic reasoning process that determines the outputs corresponding to fuzzified inputs.

The fuzzy rule-base is composed by IF-THEN rules like

$R^{(l)}$: IF ($x_1$ is $A_1^{(l)}$ and $x_2$ is $A_2^{(l)}$ and... $x_m$ is $A_m^{(l)}$) THEN ($y$ is $\omega^{(l)}$),

where: $R^{(l)}$ is the $l$th rule with $1 \leq l \leq c$ determining the total number of rules; $x_1, x_2, \ldots x_m$ and $y$ are, respectively, the input and output system variables; $A_j^{(l)}$ are the antecedent linguistic terms (or fuzzy sets) in rule ($l$) with $1 \leq j \leq m$ being the number of antecedent variables; and $\omega^{(l)}$ is the rule conclusion being, for that type of rules, a real number usually called *fuzzy singleton*. The conclusion, a numerical value and not a fuzzy set, can be considered as a *pre-defuzzified* output that helps to accelerate the inference process.

Each IF-THEN rule defines a fuzzy implication between condition and conclusion rule parts and denoted by expression $A_1^{(l)} \times \ldots \times A_m^{(l)} \to \omega^{(l)}$. The implication operator used in this work is the *product-operator*, as shown in expression (2). The right-hand term $\mu_{A_1^{(l)} \times ... \times A_m^{(l)}}(x')$ represents the condition degree and is defined in Equation (3).

$$\mu_{A_1^{(l)} \times ... \times A_m^{(l)} \to \omega^{(l)}}(x', y') = \mu_{A_1^{(l)} \times ... \times A_m^{(l)}}(x'_1, ..., x'_m) \cdot \mu_{\omega^{(l)}}(y') \qquad (2)$$

The symbol " ∗ " in Equation (3) is the *t*-norm corresponding to the conjunction *and* in the rules. The most commonly used *t*-norms between linguistic expressions *u and v* are: fuzzy intersection defined by operation *min(u,v)*, algebraic product *uv*, and the bounded sum *max(0, u+v-1)*. This work uses algebraic product as the *t*-norm operator.

$$\mu_{A_1^{(l)} \times ... \times A_m^{(l)}}(x') = \mu_{A_1^{(l)}}(x'_1) * ... * \mu_{A_m^{(l)}}(x'_m) \qquad (3)$$

Since **the rule conclusion** $\omega^{(l)}$ is considered a fuzzy singleton, the value of its membership degree $\mu_{\omega^{(l)}}(y')$ in expression (2) stays 1.0. So, the final expression for fuzzy implication degree (2) results in multiplication of each condition membership degree (3) and equal to expression (4).

$$\mu_{A_1^{(l)} \times ... \times A_m^{(l)} \to \omega^{(l)}}(x') = \mu_{A_1^{(l)} \times ... \times A_m^{(l)}}(x') \cdot (1.0) = \prod_{j=1}^{m} \mu_{A_j^{(l)}}(x'_j) \qquad (4)$$

For this type of fuzzy system, the *product inference* in Equation (3) expresses the activation degree of each identified rule by measured condition variables, and equals the expression for implication degree in (4).

The reasoning process combines all rule contributions $\omega^{(l)}$ using the centroid defuzzification formula in a weighted form, as indicated by inference function (5). This equation maps input process states $(x'_j)$ to the value resulted from inference function $Y(x')$. If we fix the structure made by the Gaussian membership functions, the parameters of the fuzzy logic system to be learned will be the rule conclusion value $\omega^{(l)}$.

$$Y(x') = \frac{\sum_{l=1}^{c} \left( \prod_{j=1}^{m} \mu_{A_j^{(l)}}(x'_j) \right) \cdot \omega^{(l)}}{\sum_{l=1}^{c} \left( \prod_{j=1}^{m} \mu_{A_j^{(l)}}(x'_j) \right)} \qquad (5)$$

## 2.3 Defuzzification

Basically, defuzzification maps output fuzzy sets defined over an output universe of discourse to crisp outputs. It is employed because in many practical applications a crisp output is required. A defuzzification strategy is aimed at producing the nonfuzzy output that best represents the possibility distribution of an inferred fuzzy output. At present, the commonly used strategies are described as the following

### 1) The Max Criterion Method

The max criterion method produces the point at which the possibility distribution of the fuzzy output reaches a maximum value.

### 2) The Mean of Maximum Method

The mean of maximum generates an output which represents the mean value of all local inferred fuzzy outputs whose membership functions reach the maximum. In the case of a discrete universe, the inferred fuzzy output may be expressed as

$$z_0 = \sum_{j=1}^{l} \frac{w_j}{l}$$

where $w_j$ is the support value at which the membership function reaches the maximum value $\mu_z(w_j)$ and $l$ is the number of such support values.

### 3) The Center of Area Method

The center of area generates the center of gravity of the possibility distribution of the inferred fuzzy output. In the case of a discrete universe, this method yields

$$z_0 = \frac{\sum_{j=1}^{n} \mu_z(w_j) w_j}{\sum_{j=1}^{n} \mu_z(w_j)}$$

where $n$ is the number of quantization levels of the output

# 3  Fuzzy Modeling

Basic principles of fuzzy models, also known in literature by *fuzzy modeling*, were first introduced by Zadeh in [2] and [13]. First applications in modeling systems using fuzzy-logic consisted initially in duplication of expert experience to process control [22]. Although, this *qualitative information* can present limitations as the acquired knowledge usually presents errors and even some gaps.

Another source of information is *quantitative information*. It is acquired by acquisition of numerical data from most representative system variables, and can be used together with the anterior qualitative information to complete it or even produce new information [3].

The acquisition of models using fuzzy logic is usually divided into two types as shown in Fig. 1: a linguistic approach composed by relational and natural models and a hybrid approach concerning the neuro-fuzzy models.

The main difference between these approaches is related to the knowledge representation in the model. While linguistic approach describes the system behaviour using rules of IF-THEN using only fuzzy sets (linguistic variables), the hybrid approach uses linguistic variables in the condition rule part (IF) and uses a numerical value in the conclusion part (THEN) which is considered as a function of input variables [3], [4].

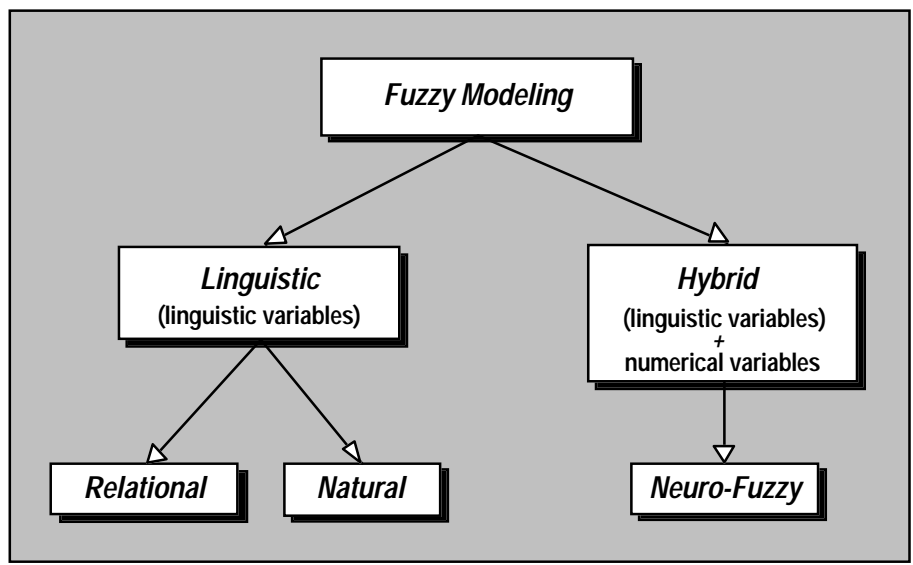

**Fig. 1: Fuzzy modeling types.**

Linguistic modeling can be divided into two types: relational modeling and *natural* modeling. Relational modeling [25-28] establishes a set of all possible rules based on an attributed linguistic partition for each input-output variable. It computes for each rule the respective true value of how much that rule contributes to describe system behaviour. The set of all rules composes, in a computational way, a multidimensional matrix called relational matrix. Using the theory of relational equations [29], [30], each matrix element can be computed as being the rule membership degree in the extracted system's model.

The second type of linguistic modeling is denoted by *natural* modeling. It does not use relational equations to obtain the model. The rules are codified from information supplied by the process operator and/or from knowledge obtained from the literature. The first application examples of this type of modeling were the fuzzy controllers in [22] and [23].

Fuzzy modeling based on hybrid approach permits employing learning techniques used by neural-networks in the identification of each rule [16], [17], [20]. The parameter set composing rule condition part are the membership functions width and their position in the respective universe of discourse. In the conclusion part, the parameters are the function terms that compute the rule answer.

# 4 The Learning Mechanism

The learning mechanism uses two data sets: one for the training stage and other to test the extracted model. Initially, using the training set, we extract the model rules and their conclusion value through a cluster-based algorithm [19]. Then, the model has its conclusion values tuned by a gradient-descent method [24] to produce the process neuro-fuzzy model. Since the test set has examples not presented during the training stage, we use it to verify the generalization model performance.

In the following subsections, we recapitulate the learning mechanism and its main characteristics.

## 4.1 Model Initialization

The first modeling stage of the electro-hydraulic actuator is concerned with the initialization of each rule conclusion using the cluster-based algorithm.

Cluster means a collection of *objects* composing a subset where its elements form a natural group among all exemplars. Therefore, it establishes a subset where the elements compose a group with common characteristics constituting a pattern. This concept applied to the fuzzy partition of system's operating domain divides it into clusters, each one interpreted as a rule $R^{(l)}$ describing, in our case, the actuator's local behavior.

The cluster concept when used with fuzzy logic [33] associates to each data point a value among zero and one representing its membership degree in the rule. This allows each sample data to belong to multiple rules with different degrees.

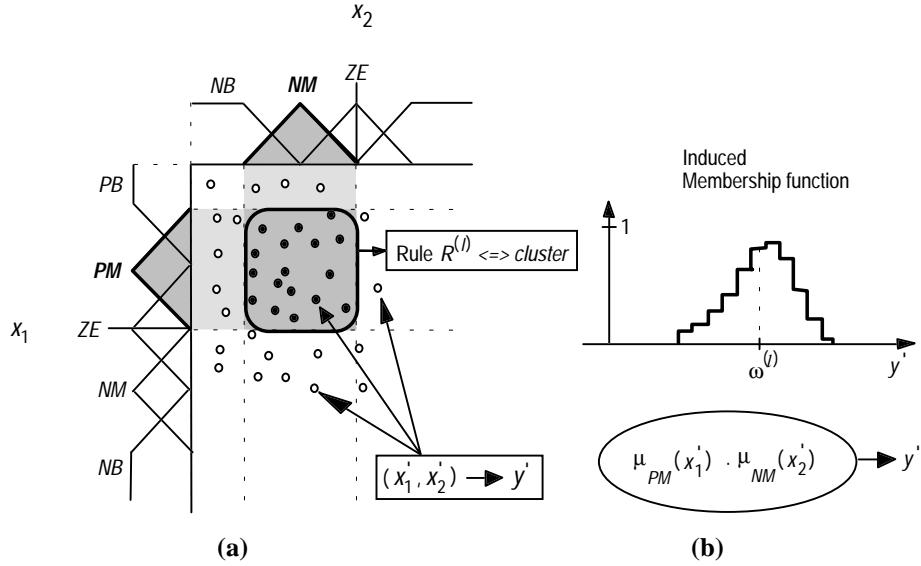

**Fig. 2: (a) Set of examples selected from the training data to extract the rule with antecedents defined by fuzzy sets *PM* and *NM*. (b) Membership function induced by weighted output values $y'$ into the specified rule region, and the computed conclusion value $\omega^{(l)}$.**

In Fig. 2, we illustrate the cluster concept applied to a fuzzy system. Suppose, for simplicity, a system with two inputs denoted by $x_1$ and $x_2$, and one output $y$. As shown in the figure, each domain variable $x_1$ and $x_2$ is equally partitioned by symmetric triangular fuzzy sets characterising each linguistic term, for example, with *PM- Positive Medium*, *NB- Negative Big*, *ZE- Zero*, and other fuzzy sets.

A data set composed by system examples is acquired to be used in the training stage. Fig. 2(a) displays the examples covering the domain, each one formed by a data sample like $(x_1', x_2') \rightarrow y'$. The examples are grouped in clusters for each respective rule $R^{(l)}$. In the figure, we exemplify the rule acquisition expressed in statement (6).

**IF** [($x_1$ is *Positive-Medium*) **and** ($x_2$ is *Negative-Medium*)]**THEN** [$y$ is $\omega^{(l)}$]   (6)

The condition rule part is characterized by fuzzy sets *PM (Positive-Medium)* and *NM (Negative-Medium)*. The conclusion part, characterised by a numerical value $\omega^{(l)}$, is extracted based on the examples contained into the domain region covered by the two fuzzy sets *PM* and *NM*. This set of examples is represented in Fig. 2 by filled circles into the rule region $R^{(l)}$.

Using the fuzzy cluster concept, it attributes to each example a certain degree of how much it belongs to that cluster or, in other words, how much each example contributes to the extraction of conclusion value $\omega^{(l)}$ of that rule $R^{(l)}$.

Suppose an example $(x_1^{'}, x_2^{'}) \rightarrow y^{'}$ inside the rule region. Its contribution degree is computed by the product of each condition membership degree in fuzzy sets *PM* and *NM* of specified rule region, as expressed in (7) and displayed in Fig. 2(b). The computed contribution degree then weights the corresponding output value $y^{'}$.

$$\mu_{PM}(x_1^{'}) \cdot \mu_{NM}(x_2^{'}) \qquad (7)$$

The anterior operations are executed for each example inside the rule region, and compose a membership function defined for all output values $y^{'}$ into the rule region, as Fig. 2(b) illustrates to rule (6). Using the centroid method, the final conclusion value $\omega^{(l)}$ for that rule ($l$) is computed from the induced membership function.

## 4.2 The Cluster-Based Algorithm

The algorithm uses the ideas introduced in the anterior section to extract each rule to build an initial model to the electro-hydraulic actuator. At first, the algorithm divides system's domain into a set of clusters using the fuzzy sets attributed to each variable. As shown in Fig. 2, each cluster represents a local rule. The rules composing the model are established *a priori* by multiplication of the number of fuzzy sets attributed to each condition variable.

The cluster-based algorithm steps are described below in more detail, and a simple example illustrates it.

Starting with rule one ($l = 1$) and the $k$th training example, the cluster-based algorithm summarizes the following steps to extract its conclusion value $\omega^{(1)}$:

**Step 1)** Establish the variable set better characterizing the actuator's behaviour;

**Step 2)** Set the limits of each universe of discourse and the number of fuzzy sets for the selected input-output variables in step 1. The algorithm uses symmetric gaussian membership functions uniformly distributed by each universe of discourse;

**Step 3)** The algorithm begins with the extraction of the first rule ($l = 1$). From the training set, we take the $k$th numerical example $(x_1^{'}(k), x_2^{'}(k), \ldots, x_m^{'}(k)) \rightarrow y^{'}(k)$, and calculate, for all condition variables, their respective membership degrees in the fuzzy sets composing the rule as expressed in (8).

$$\mu_{A_1^{(l)}}(x_1^{'}(k)), \mu_{A_2^{(l)}}(x_2^{'}(k)), \ldots, \mu_{A_m^{(l)}}(x_m^{'}(k)) \qquad (8)$$

**Step 4)** Calculate the membership degree of corresponding output value $y^{'}(k)$ in rule ($l$), or its membership degree in cluster ($l$), as indicated in (9) by the term $S1^{(l)}(k)$.

$$S1^{(l)}(k) = \mu_{A_1^{(l)}}(x_1^{'}(k)) \cdot \mu_{A_2^{(l)}}(x_2^{'}(k)) \cdots \mu_{A_m^{(l)}}(x_m^{'}(k)) \qquad (9)$$

**Step 5)** The output value $y^{'}(k)$ is weighted by its membership degree $S1^{(l)}(k)$ in rule ($l$), as described in Equation (10) by $S2^{(l)}(k)$.

$$S2^{(l)}(k) = y^{'}(k) \cdot S1^{(l)}(k) \qquad (10)$$

**Step 6)** In this step, the algorithm adds recursively the value $S2^{(l)}(k)$ and the membership degree $S1^{(l)}(k)$ as indicated in (11). The variable *Numerator* adds to rule ($l$) all weighted contributions made by the $n$ data values $y^{'}(k)$ in the training set. The variable *Denominator* sums each membership degree in order to normalize the conclusion value $\omega^{(l)}$.

$$\begin{cases} Numerator \rightarrow Numerator + (S2^{(l)}(k)) \\ Denominator \rightarrow Denominator + (S1^{(l)}(k)) \end{cases} \qquad (11)$$

Get the next example. If there are no more examples, go to step 7 and compute the conclusion value $\omega^{(l)}$. If not, go to step 3 and pick up the next example as indicated in (12).

$$k \rightarrow k+1 \qquad (12)$$

**Step 7)** If the training set has finished ($k = n$), compute the conclusion value $\omega^{(l)}$ for rule ($l$) using equation (13).

$$\omega^{(l)} = \frac{Numerator}{Denominator} \qquad (13)$$

**Step 8)** The algorithm now goes to next rule (14), begins again with the first training example (15), and returns to step 3. If there are no more rules ($l = c$), the algorithm stops.

$$l \rightarrow l+1 \qquad (14)$$

$$k \rightarrow 1 \qquad (15)$$

## 4.3 Illustrative Example

This example illustrates the anterior steps for one training period. It uses the two examples shown in (16) to demonstrate the computation of $\omega^{(l)}$ for a certain specified rule ($l$).

$$\begin{cases} (x_1'(1), x_2'(1)) \to y'(1) & (k=1) \\ (x_1'(2), x_2'(2)) \to y'(2) & (k=2) \end{cases} \quad (16)$$

This example considers a system with two antecedent variables denoted by $x_1$ and $x_2$, and one output variable, $y$. The variables are partitioned by symmetric triangular membership functions. The use of a triangular partition instead of a gaussian one helps us to better visualize the algorithm steps. We attributed 7 fuzzy sets to variable $x_1$ (Fig. 3a), 5 fuzzy sets to $x_2$ (Fig. 3b), and 5 fuzzy sets to $y$ (Fig. 3c).

Suppose in this example that we want to extract the consequent value $\omega^{(1)}$ for rule ($l = 1$) described in (17).

$$R^{(1)}(k): \textbf{IF } (x_1 \text{ is } A_3^{(1)} \textbf{ and } x_2 \text{ is } A_4^{(1)}) \textbf{ THEN } \omega^{(1)} = ? \quad (17)$$

Each variable in the two training samples in (16) has a membership degree in each antecedent fuzzy set $A_3^{(1)}$ and $A_4^{(1)}$. In expressions (18) to (20), we show the corresponding degrees attributed to values $x_1'$, $x_2'$, and $y'$, for examples in (16). The triangular partition causes all numerical values to always have two non-zero membership degrees and a null degree in the other fuzzy sets, as illustrated in Fig. 3 for each training example. The difference using gaussian functions is that each variable would have a number of degrees equal to the attributed fuzzy sets.

$$\begin{cases} x_1'(1) \to \left[ 0, \mu_{A_2^{(1)}}(x_1'(1)), \mu_{A_3^{(1)}}(x_1'(1)), 0, 0, 0, 0 \right] \\ x_1'(2) \to \left[ 0, \mu_{A_2^{(1)}}(x_2'(2)), \mu_{A_3^{(1)}}(x_2'(2)), 0, 0, 0, 0 \right] \end{cases} \quad (18)$$

$$\begin{cases} x_2'(1) \to \left[ 0, 0, \mu_{A_3^{(1)}}(x_2'(1)), \mu_{A_4^{(1)}}(x_2'(1)), 0 \right] \\ x_2'(2) \to \left[ 0, 0, \mu_{A_3^{(1)}}(x_2'(2)), \mu_{A_4^{(1)}}(x_2'(2)), 0 \right] \end{cases} \quad (19)$$

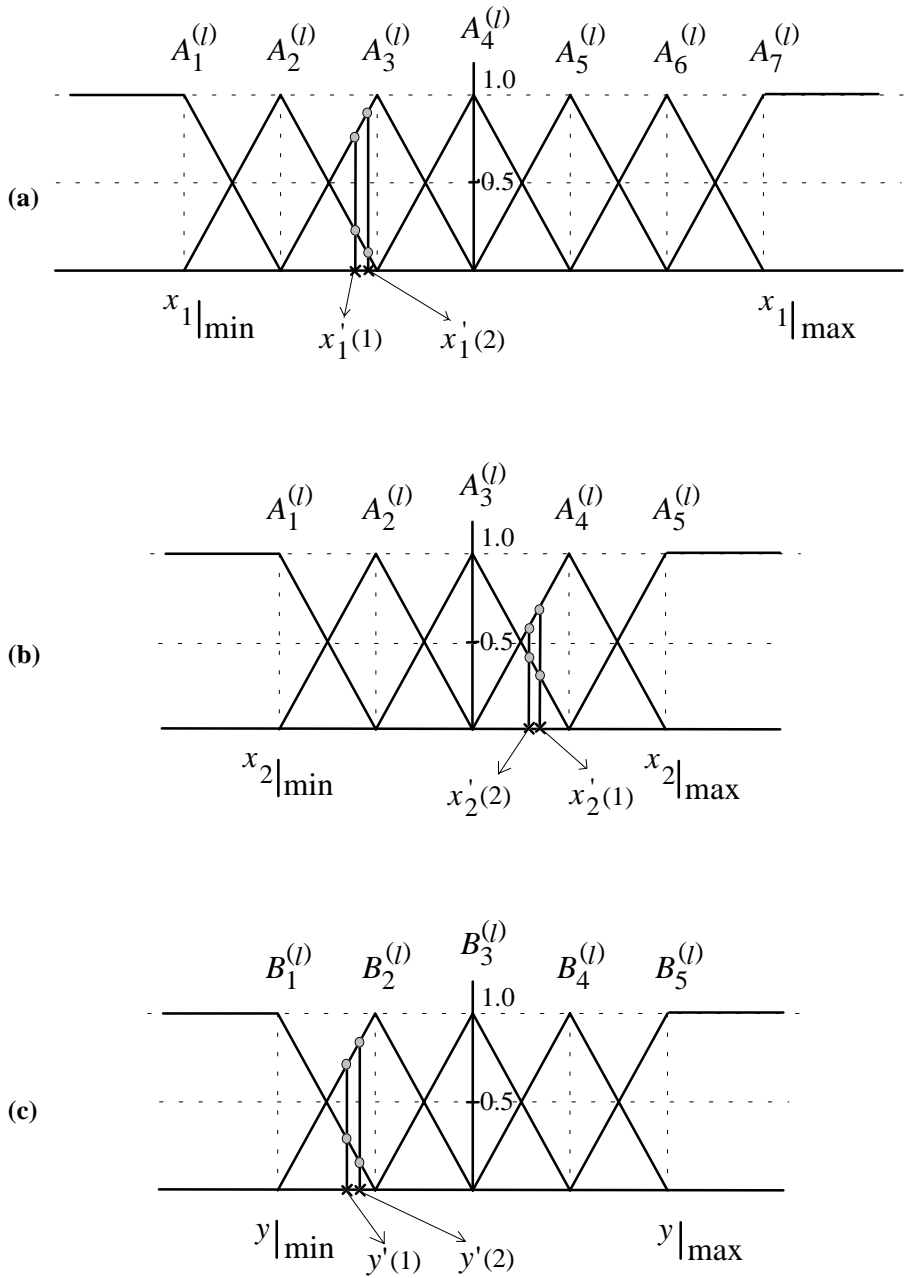

**Fig. 3: (a)** Partition of variable $x_1$ with 7 fuzzy sets. **(b)** Partition of variable $x_2$ with 5 fuzzy sets. **(c)** Partition of variable $y$ with 5 fuzzy sets.

$$\begin{cases} y'(1) \rightarrow \left[ \mu_{B_1^{(1)}}(y'(1)), \mu_{B_2^{(1)}}(y'(1)), 0, 0, 0 \right] \\ y'(2) \rightarrow \left[ \mu_{B_1^{(1)}}(y'(2)), \mu_{B_2^{(1)}}(y'(2)), 0, 0, 0 \right] \end{cases} \quad (20)$$

The algorithm extracts the value of $\omega^{(1)}$ using steps 5 to 8. It considers the fuzzy sets $A_3^{(l)}$ and $A_4^{(l)}$ of condition part in rule $R^{(1)}$. Therefore, the conclusion value is computed by Equation (21) combining each output value $y'(k)$, weighted and normalized by their contribution degrees to the specified rule.

$$\omega^{(1)} = \frac{\left[ \mu_{A_3^{(l)}}(x_1'(1)) \cdot \mu_{A_4^{(l)}}(x_2'(1)) \right] \cdot y'(1) + \left[ \mu_{A_3^{(l)}}(x_1'(2)) \cdot \mu_{A_4^{(l)}}(x_2'(2)) \right] \cdot y'(2)}{\left[ \mu_{A_3^{(l)}}(x_1'(1)) \cdot \mu_{A_4^{(l)}}(x_2'(1)) \right] + \left[ \mu_{A_3^{(l)}}(x_1'(2)) \cdot \mu_{A_4^{(l)}}(x_2'(2)) \right]} \quad (21)$$

## 4.4 The Neuro-Fuzzy Algorithm

The neuro-fuzzy algorithm developed by Wang [24] uses the hybrid model developed by Takagi-Sugeno in [3]. In this type of model, condition part uses linguistic variables and the conclusion part is represented by a numerical value which is considered a function of system's condition expressed in the variables $x_1, x_2, \ldots, x_m$ (22). These models are suitable for neural-based-learning techniques as gradient methods to extract the rules [6] and generate models with a reduced number of rules.

$$\omega^{(l)} = g(x_1, x_2, \ldots, x_m) \quad (22)$$

The neuro-fuzzy algorithm uses membership functions of gaussian type. With gaussian fuzzy sets, the algorithm is capable of utilizing all information contained in the training set to calculate each rule conclusion, which is different when using triangular partitions.

Fig. 4 illustrates the neuro-fuzzy scheme for an example with two input variables $(x_1, x_2)$ and one output variable ($y$). In the first stage of the neuro-fuzzy scheme, the two inputs are codified into linguistic values by the set of gaussian membership functions attributed to each variable. The second stage calculates to each rule $R^{(l)}$ its respective activation degree. Last, the inference mechanism weights each rule conclusion $\omega^{(l)}$, initialized by the cluster-based algorithm, using the activation degree computed in the second stage. The error signal between the model inferred value $Y$ and the respective measured value (or teaching value) $y'$, is used by the gradient-descent method to adjust each rule conclusion. The algorithm changes the values of $\omega^{(l)}$ to minimize an objective function $E$ usually expressed by the mean quadratic error (23). In this equation, the value $y'(k)$ is the desired output value related with the condition vector $x'(k) = (x'_1, x'_2, \cdots, x'_m)_k$. The element $Y(x'(k))$ is the inferred response to the same condition vector $x'(k)$ and computed by Equation (24).

$$E = \frac{1}{2}\left[Y(x'(k)) - y'(k)\right]^2 \qquad (23)$$

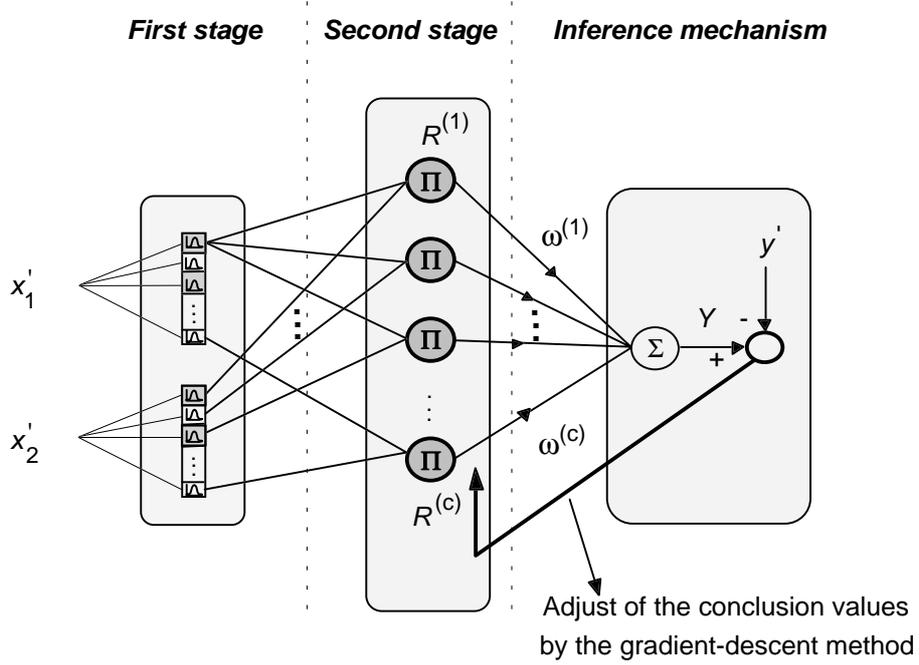

**Fig. 4: The neuro-fuzzy scheme.**

$$Y(x'(k)) = \frac{\sum_{l=1}^{c}\left(\prod_{j=1}^{m}\mu_{A_j^{(l)}}(x'_j(k))\right)\omega^{(l)}(k)}{\sum_{l=1}^{c}\left(\prod_{j=1}^{m}\mu_{A_j^{(l)}}(x'_j(k))\right)} \qquad (24)$$

Equation (25) establishes adjustment of each conclusion $\omega^{(l)}$ by the gradient-descent method. The symbol $\alpha$ is the learning rate parameter, and $t$ indicates the number of learning iterations executed by the algorithm.

$$\omega^{(l)}(t+1) = \omega^{(l)}(t) - \alpha \frac{\partial E}{\partial \omega^{(l)}} \tag{25}$$

The inference function (24) depends on $\omega^{(l)}$ only through its numerator. The expression composing the numerator is now denoted by $a$ and is shown in (26).

$$a = \sum_{l=1}^{c} \left( \prod_{j=1}^{m} \mu_{A_j^{(l)}}(x_j^{'}(k)) \right) \omega^{(l)}(t) \tag{26}$$

The denominator of function (24) is dependent on a term $d^{(l)}$, defined in (27), and denoted by $b$ in (28).

$$d^{(l)} = \prod_{j=1}^{m} \mu_{A_j^{(l)}}(x_j^{'}(k)) \tag{27}$$

$$b = \sum_{l=1}^{c} (d^{(l)}) \tag{28}$$

To calculate the adjustment of each conclusion value $\omega^{(l)}$, it is necessary to compute the variation of the objective function $E$, $\partial E$, in relation to the variation that occurred in $\omega^{(l)}$ in the anterior instant, $\partial \omega^{(l)}$. Therefore, using the chain rule to calculate $\partial E / \partial \omega^{(l)}$ results in expression (29).

$$\frac{\partial E}{\partial \omega^{(l)}} = \frac{\partial E}{\partial Y} \frac{\partial Y}{\partial a} \frac{\partial a}{\partial \omega^{(l)}} \tag{29}$$

The use of chain rule looks for the term contained in $E$ that is directly dependent on the value to be adjusted, i.e., the conclusion value $\omega^{(l)}$. Therefore, we can verify by chain equation (29) that it starts with $E$ dependent of $Y$ value, the $Y$ value depends on term $a$ and, at last, the expression $a$ is a function of $\omega^{(l)}$.

Using Equations (26) to (28), the $Y$ function is written as (30).

$$Y(x^{'}(k)) = \frac{a}{b} \tag{30}$$

The three partial derivatives of chain rule are computed resulting in equations (31), (32), and (33).

$$\frac{\partial E}{\partial Y} = (Y(x^{'}(k)) - y^{'}(k)) \tag{31}$$

$$\frac{\partial Y}{\partial a} = \frac{1}{b} \tag{32}$$

$$\frac{\partial a}{\partial \omega^{(l)}} = \prod_{j=1}^{m} \mu_{A_j^{(l)}}(x_j^{'}(k)) = d^{(l)} \tag{33}$$

Substituting the three derivatives in chain equation (29), the final partial derivative of $E$ in relation to $\omega^{(l)}$ results in expression (34).

$$\frac{\partial E}{\partial \omega^{(l)}} = \frac{(Y(\mathbf{x}'(k)) - y'(k))d^{(l)}}{b} \tag{34}$$

The replacement of derivative $\partial E / \partial \omega^{(l)}$ in Equation (25) gives the final result presented in (35). In this equation, $d^{(l)}$ represents the activation degree of rule ($l$) by condition $\mathbf{x}'(k)$. The expression $\sum_{l=1}^{c}(d^{(l)})$ is the normalization factor of value $d^{(l)}$. Using these two considerations, the adjustment to be made in $\omega^{(l)}$ can be interpreted as being proportional to the error between the neuro-fuzzy model response and the supervising value, but weighted by the contribution of rule ($l$), denoted by $d^{(l)}$, to the final neuro-fuzzy inference.

$$\begin{aligned} \omega^{(l)}(t+1) &= \omega^{(l)}(t) - \alpha \frac{(Y(\mathbf{x}'(k)) - y'(k))d^{(l)}}{b} \\ &= \omega^{(l)}(t) - \alpha \frac{(Y(\mathbf{x}'(k)) - y'(k))d^{(l)}}{\sum_{l=1}^{c}(d^{(l)})} \end{aligned} \tag{35}$$

# 5 The Experimental System

The experimental control system is composed by a permanent-magnet (P.M.) synchronous motor driving a hydraulic pump that sends fluid to move a linear piston. Fig. 5 shows a diagram of the system incorporating two control loops. The interior loop, in grey, is responsible for the motor speed control. The loop is composed of an electrical drive with a PI controller to command the motor speed. The exterior loop, in black, controls the piston position using a proportional controller that gives the motor speed reference to the electrical drive.

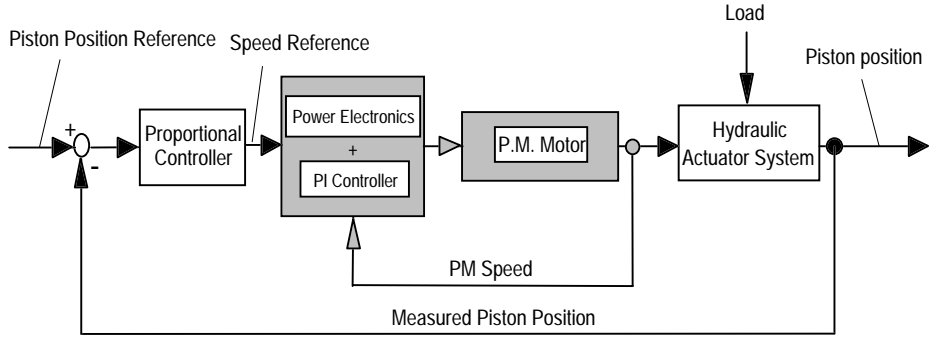

**Fig. 5: Diagram of the experimental electro-hydraulic drive system.**

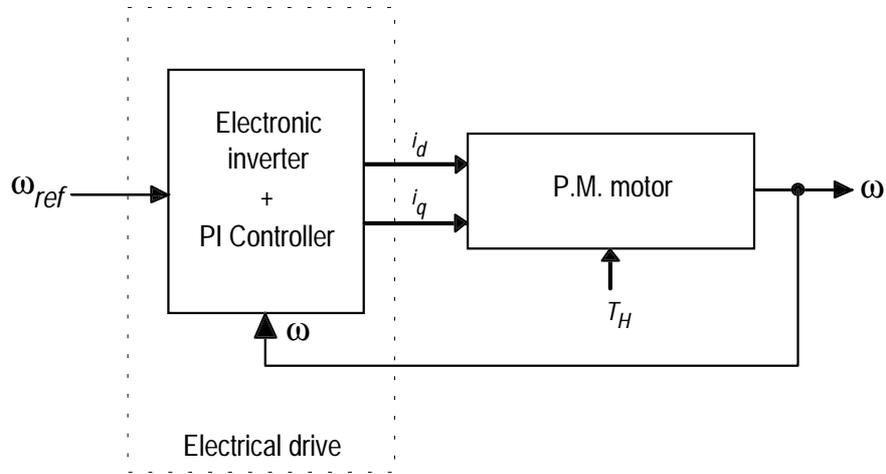

**Fig. 6: First subsystem composed by the electrical drive and the P.M. motor.**

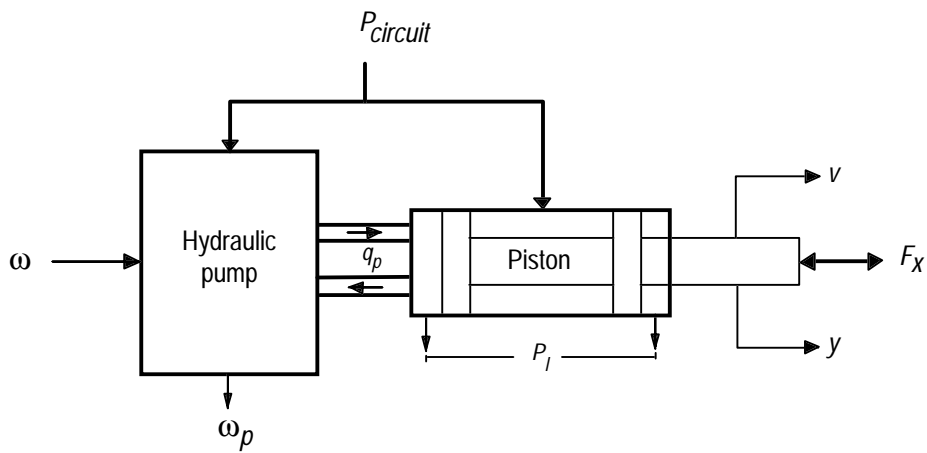

**Fig. 7: Second subsystem composed by the hydraulic system.**

Two subsystems compose the actuator. Figures 6 and 7 show these subsystems. The first subsystem in Fig. 6 shows the electrical drive that controls the motor speed ($\omega$). The electronic inverter employs IGBTs to generate currents $i_1, i_2, i_3$, in *Park* coordinates $i_d$ and $i_q$ as shown in the figure, commanding the P.M. motor (220V/ 1.2Nm/ ±3000 rpm). The speed controller is composed of a PI regulator. The motor load is denoted by $T_H$, and it comes from the hydraulic pump connected to the motor.

In Fig. 7, we show the second subsystem that composes the electro-hydraulic actuator. The hydraulic pump is assumed to rotate at the same speed as the motor ($\omega = \omega_p$), with the hydraulic circuit operating at a pressure of 40 bar ($P_{circuit} = 40 bar$). As the pump sends fluid ($q_p$) to the piston, the pressure difference ($P_l$) in the piston induces a force that moves it. The implemented experimental system permits connection of an inertial variable load to the piston represented in the figure by the symbol $F_x$.

The electro-hydraulic system is marked by a nonlinear characteristic localized into the hydraulic circuit dominating its behaviour. This characteristic introduces a non-linear interface between the electrical system and the hydraulic actuator. In Fig. 8, we display an experimental curve illustrating the relationship between pump speed signal ($\omega$), considered equal to the motor speed, and the piston linear speed ($v$) which is associated with the fluid quantity $q_p$ sent by the pump. The curve shows an asymmetric dead-zone localized between the pump speed values of -700 r.p.m. and +900 r.p.m., and it displays a hysteresis effect out of the dead-zone. When operating into the dead-zone, the two actuator subsystems stay disconnected and the piston stops as the fluid stream $q_p$ debited by the pump is near zero. Out of the dead-zone, the inclination of the two lines shows that the pump debits slightly more hydraulic fluid when rotating in one direction than rotating to the other.

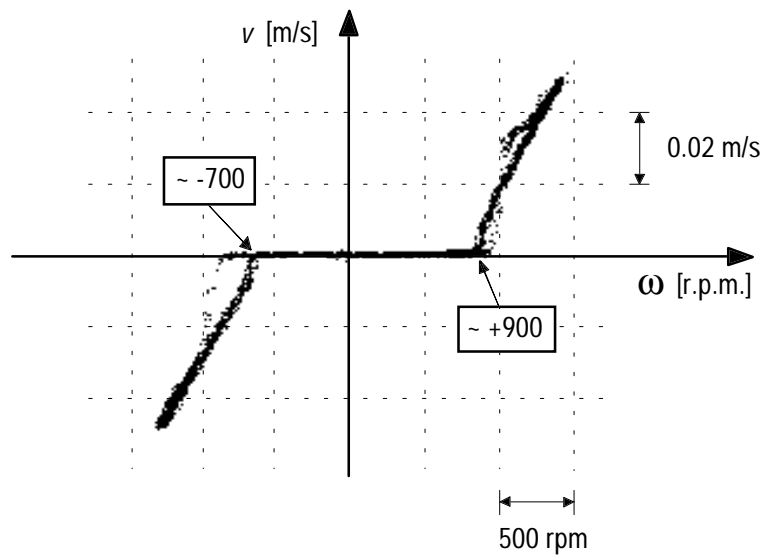

**Fig. 8: Experimental curve showing the nonlinear characteristic present in the hydraulic circuit.**

In Fig. 9, we illustrate the piston asymmetric behavior when operating in open-loop (without the proportional controller) for a sinusoidal reference to the motor speed (Fig. 9a). We can observe in Fig. 9b that the piston moves more to one direction than to the other. Therefore, after some sinusoidal periods, the piston halts at the end of its course of 0.20m. This behavior is mainly caused by the nonlinear characteristic with the asymmetry of the dead-zone, sending more fluid for one pump speed direction than to the other.

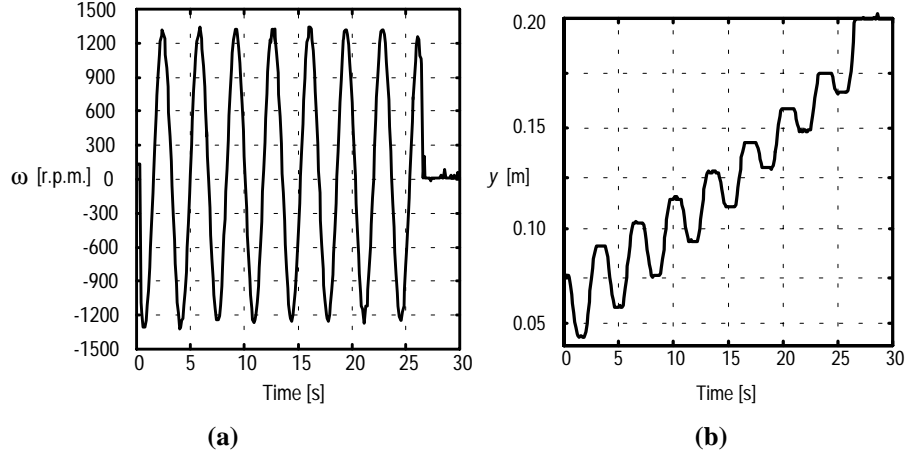

**Fig. 9: Actuator's response for a sinusoidal reference signal with amplitude and frequency constants, operating in an open-loop mode. (a) Motor speed signal ($\omega$). (b) Piston position signal ($y$).**

## 5.1 Training Data Generation

To obtain some relevant information for the training process, we used theoretical knowledge about system physics. This knowledge is present when we model the actuator using electromechanical power conversion theory and hydrodynamic laws. As the system contains a great number of variables that can be chosen to characterize its dynamic, it is important to make some hypotheses and simplifications to concentrate our attention to a small but representative variable set.

As shown before, the electro-hydraulic actuator is separated into two subsystems: the electrical drive and the hydraulic circuit with the pump and piston elements. If we consider these subsystems as "black-boxes" and make some considerations, as, for example, not consider relevant the contribution of the pressure signal in the circuit ($P_{circuit}$) because it remains approximately constant during actuator's operation, we can interpret the piston position signal ($y$) as a function of the reference signal ($y_{ref}$), the motor speed ($\omega$), and the linear speed of the piston ($v$). Thus, the direct model can be represented by relation (36).

$$y = f(y_{ref}, \omega, v) \qquad (36)$$

To extract function $f(.)$, it is necessary to use some numerical data available from the system. For this, two different sets of experimental values are added to the modeling process, one set for training and the other for testing.

As described, the actuator has an asymmetric behavior dominated by the presence of a nonlinear characteristic. If we need to acquire some training data that characterizes a significant part of the electro-hydraulic system's operating domain, we cannot use the system in an open-loop mode (see Fig. 9) since we cannot control the system. So, to assure that the training data contain representative data and attenuate the nonlinear characteristic effects, we used the actuator in a closed-loop with a proportional regulator for a coarse piston position control.

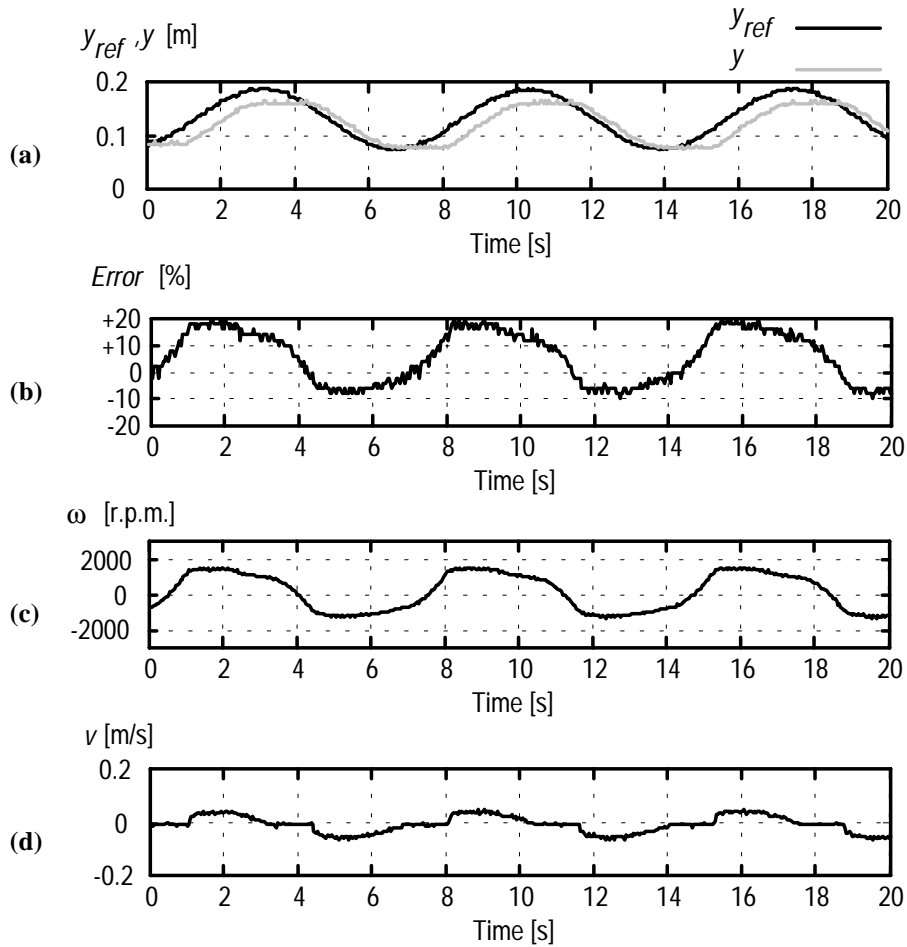

**Fig. 10: Electro-hydraulic system behaviour when operating with the closed-loop proportional controller. (a) Reference signal evolution ($y_{ref}$) and the piston position signal ($y$). (b) Error signal evolution displayed in a percentage scale. (c) Evolution of the pump speed signal ($\omega$). (d) Evolution of the piston speed signal ($v$).**

In Fig. 10, we show the actuator's evolution when it operates with the proportional closed-loop controller under a sinusoidal reference signal. The use of a coarse controller as the proportional one helps us to accent the highly nonlinear character of the actuator.

As Fig. 10a shows, the piston follows its reference signal with an asymmetric time-delay causing high tracking errors. As the pump dead-zone is large for positive speeds, there is a larger delay in the system's response resulting in high errors (Fig. 10b). On the contrary, as negative dead-zone is shorter, the system responds faster and the error signal decreases.

If we link the pump speed signal displayed in Fig. 10c with the respective piston speed signal in Fig. 10d, we can note that there is a set of operating regions where, although the pump rotates, the piston does not move. Fig. 11 shows a zoom of this behavior. For the pump speed signal, we mark the speed interval corresponding to the dead-zone. Below, we mark the corresponding regions where the piston speed is zero. When the pump operates into the dead-zone, the hydraulic circuit is decoupled from the electrical part. The pump, although rotating, does not debit fluid into the hydraulic system and so there is no pressure difference on the piston to move it.

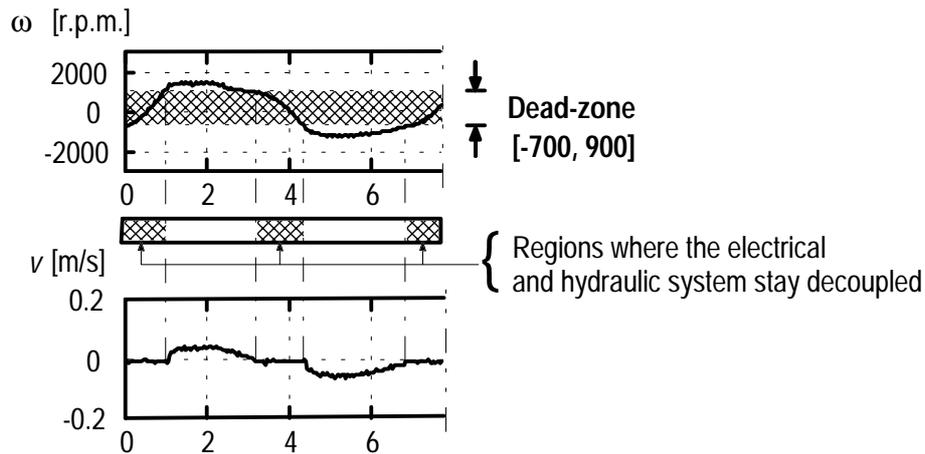

**Fig. 11: Picture detail of the pump speed and piston speed signals. It shows the effect of the dead-zone decoupling the hydraulic part from the electrical one.**

To complement the theoretical knowledge about the experimental system with more objective information, some experimental data is acquired. This data set is used in the training stage and is composed of the system's behavior examples.

Usually, to construct a training set, a Pseudo-Random Binary Signal (PRBS) is injected into the system in the manner that collected data spans during system's operating domain, although, this signal is not good to excite drive systems as pointed out in [7]. So, a better technique is to use an excitation signal of sinusoidal type composed of different magnitudes and frequencies, but within drive's response limits.

For the electro-hydraulic actuator, we used a sinusoidal signal as the reference for piston position with its amplitude ranging from 0 to 0.2m (the piston course limits) and frequencies among 0 and 1Hz because, for higher frequency values, the actuator begins filtering the reference signal.

The modeling process is described by a diagram in Fig. 12. Initially, a data set with four system signals ($y_{ref}, \omega, v, y$) is acquired using the anterior training procedures. Fig. 13 displays the acquired training set composed of the sinusoidal reference signal $y_{ref}$ with respective position $y$, the hydraulic pump speed signal $\omega$, and the piston speed signal, $v$.

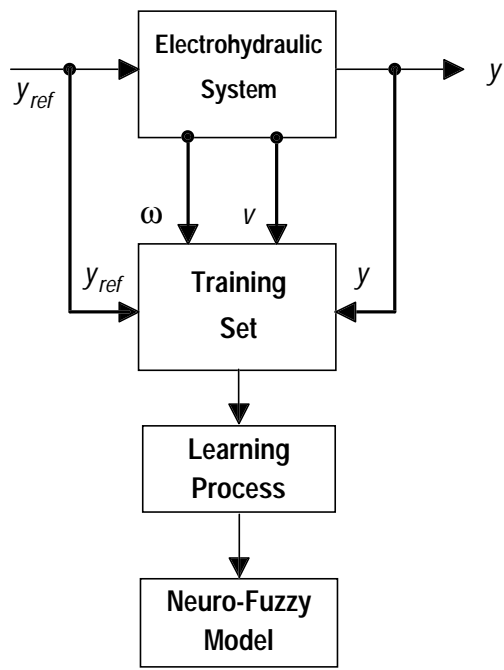

**Fig. 12: Diagram scheme representing the modeling stages.**

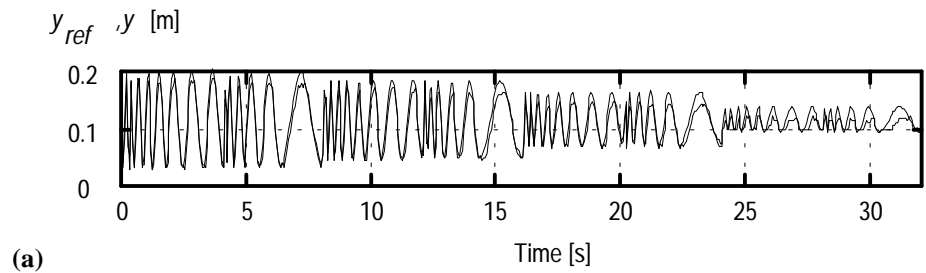

(a)

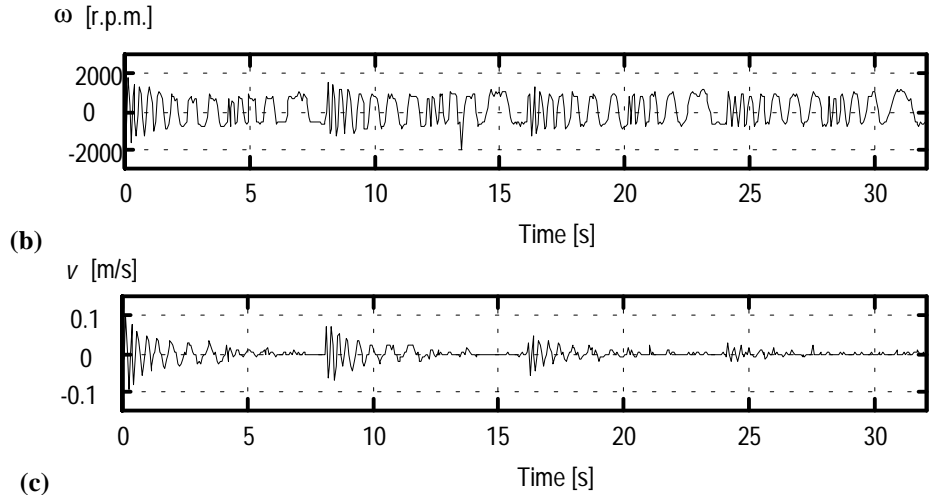

Fig. 13: The acquired training data set. (a) Reference and position signal ($y_{ref}$ and $y$). (b) Hydraulic pump speed ($\omega$). (c) Piston speed ($v$).

# 6 Neuro-Fuzzy Modeling of The Electro-Hydraulic Actuator

In this section, the actuator is modelled using the neuro-fuzzy algorithm based on training data set of Fig. 13. The experiment consists of obtaining the inverse model of the actuator represented by relation $\omega = h(y_{ref}, y, v)$.

The fuzzy model is composed of 7 membership functions attributed to the reference signal $y_{ref}$, 11 membership functions to the piston position signal $y$, and 7 membership functions attributed to the piston speed $v$. The functions are of gaussian type, as explained before, with their shape $b_j^i$ settled in 60% of each partition interval for each variable ($j$).

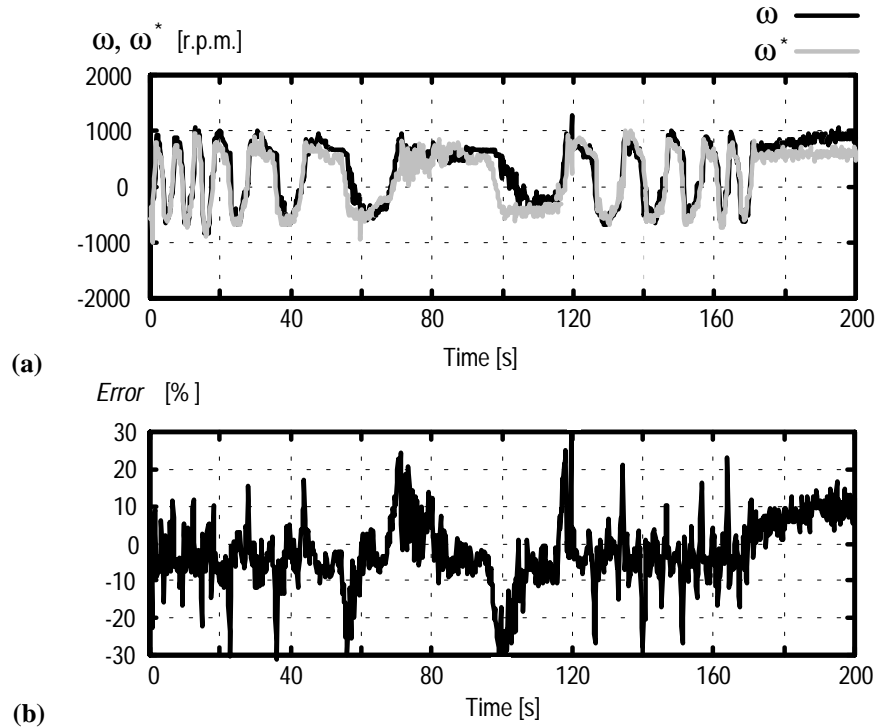

**Fig. 14: Modeling results obtained using the cluster-based algorithm to extract the initial actuator's fuzzy inverse-model. (a) Evolution of the measured ($\omega$) and the inferred pump speed signal ($\omega^*$). (b) Error signal evolution.**

The first step of modeling process uses the cluster-based algorithm to extract the initial fuzzy model. To verify the generalization capability of the learned model, we use a test data set with actuator's examples not presented to the learning algorithm during the training stage. Fig. 14 displays the generalization results obtained after extracting the fuzzy inverse-model.

Fig. 14a shows the inferred pump speed ($\omega^*$) from the fuzzy model and the measured one ($\omega$). Through the error signal displayed in Fig. 14b, we can observe that there are high errors for some operating regions. These are caused mainly by

- those domain regions where a small number or even no examples were acquired, because there was not enough information to extract a representative rule set for those regions;

- when the actuator operates into the dead-zone, it cannot be defined an inverse functional relation and the model generates high prediction errors;

- other errors appear as a consequence of noise presence in acquired signals *y* and *v*, which can deviate the inferred pump speed values from their correct predictions within a certain degree.

In the next experiment, we consider the anterior initial model and the use of the gradient-descent method explained in Section 4.4 to fine adjust it. For the learning process, the parameters used by the algorithm were: a number of 50 iterations ($K = 50$), a learning rate of 0.8 ($\alpha = 0.8$), and the same fuzzy model structure used in the cluster-based algorithm. The results obtained are displayed in Fig. 15. They show the good tuning made by the neuro-fuzzy algorithm that reduces the error signal to about 10%.

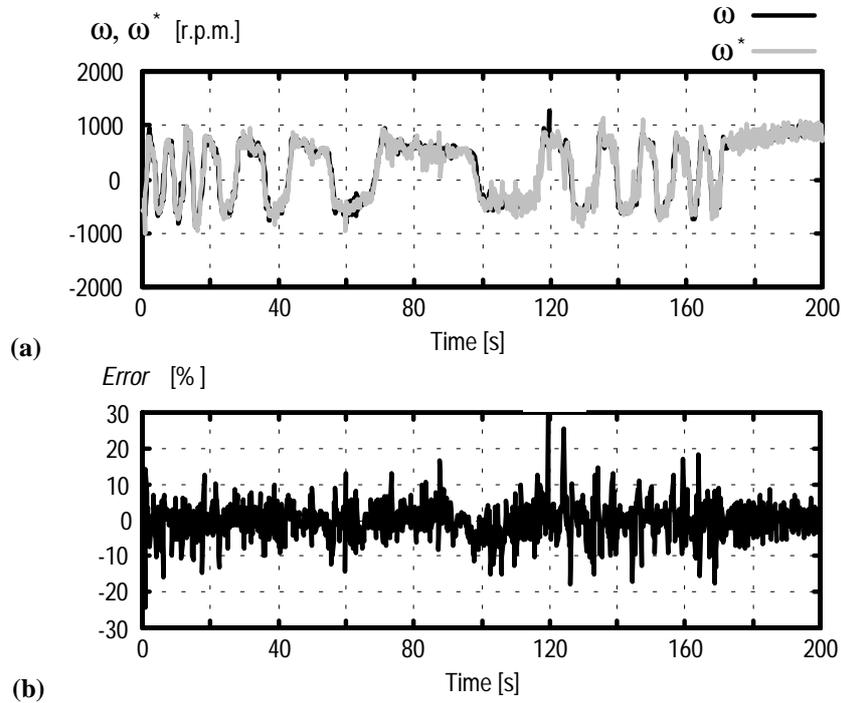

**(a)**

**(b)**

**Fig. 15: Modeling results obtained after tuning the initial model using the neuro-fuzzy algorithm. (a) Evolution of the measured ($\omega$) and the inferred signal ($\omega^*$). (b) Error signal evolution.**

It is important to note that the number of iterations and the learning rate were chosen in a manner that prohibits the model from incorporating noise dynamics by overfitting. Another aspect of domain regions where the number of acquired examples is small, the neuro-fuzzy algorithm continues to present high inference errors because it had little or no information to do a good tuning and extract representative rules. These points reveal the necessity of acquiring real-time information from the process. In this way, the learning mechanism can collect more information to correct and/or incorporate other rules into the model and reduce its prediction errors.

# 7 The Neuro-Fuzzy Control System

This section describes the neuro-fuzzy control system and shows experimental results of the electro-hydraulic position control. In the neuro-fuzzy control system, which is based on the feedback-error-learning scheme, each rule conclusion $\omega^{(l)}$ is modified by the gradient-descent method to minimize the mean quadratic error $E$. In the implemented controller, the neuro-fuzzy model minimizes the mean quadratic error generated by the proportional controller (P) to adjust each rule as indicated in Equation set (37).

$$\left[ \begin{array}{l} E = \dfrac{1}{2}\left(\mathrm{P}(y - y_{ref})\right)^2 \\ \omega^{(l)}(t+1) = \omega^{(l)}(t) - \alpha \dfrac{\partial E}{\partial \omega^{(l)}} \end{array} \right. \qquad (37)$$

Fig. 16 shows a diagram of our control scheme. The control system operates in two levels. The *high level* contains the responsible learning mechanism by actualization of the information contained in the inverse relation. The *low level* constitutes the control system formed by the feedback-loop and a feedforward-loop composed by the fuzzy inverse relation $\omega_{comp} = h(y_{ref}, v, y)$ with its inference mechanism producing the compensation signal $\omega_{comp}$.

At each control iteration, the learning system collects the present values of the reference signal ($y_{ref}$), piston speed ($v$), and the current piston position ($y$), through the available sensor set. These signals express actuator's operating condition and make each model rule active to some degree (see expression (3)). The inference mechanism uses the model rules with corresponding activation degrees and computes the compensation signal ($\omega_{comp}$) to be added to the proportional controller command ($\omega_p$), as illustrated in Fig. 16. The final signal, denoted by $\omega_{ref}$ and equal to the sum of $\omega_{comp}$ and $\omega_p$ ($\omega_{ref} = \omega_p + \omega_{comp}$), is sent to the electro-hydraulic actuator as its command signal.

To conclude the control cycle, the error signal generated by the proportional controller after the application of computed compensation signal is used to adjust each rule. The inverse relation is then adjusted based on the performance attained by the compensation made with the anterior rule set and verified through the magnitude of the proportional controller signal. Each rule is then adjusted proportionally to its anterior activation degree, interpreted as a measure of how much the rule contributed to the actuator's actual performance.

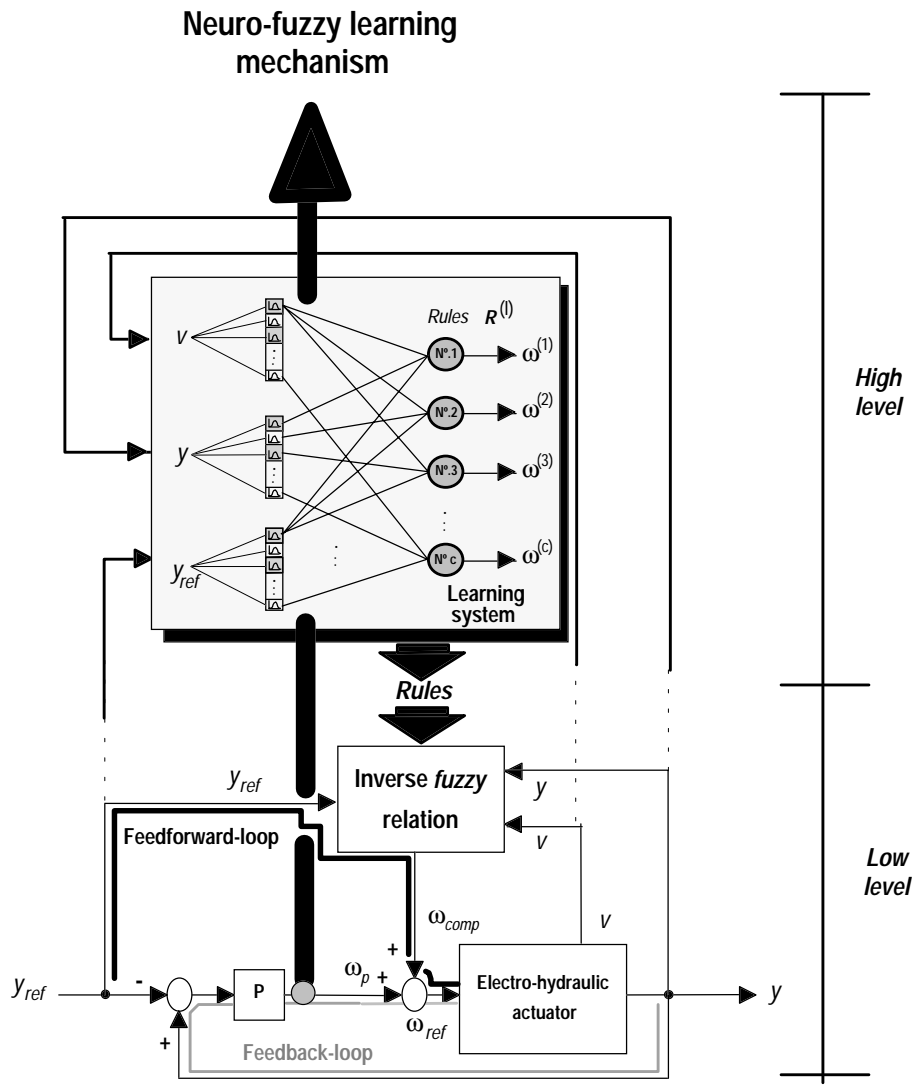

**Fig. 16: Diagram of the implemented neuro-fuzzy control system.**

## 7.1 Experimental Results

The experimental results use a square wave as the reference position signal to the piston. Fig. 17 shows the results of the first test. In this test, the actuator is controlled only through the feedback-loop with the proportional controller without any compensation signal. The results show an offset error signal between the reference position and that attained by the piston (Fig. 17a). The asymmetric error evolution shown in Fig. 17b is conditioned by the asymmetric dead-zone in the hydraulic circuit.

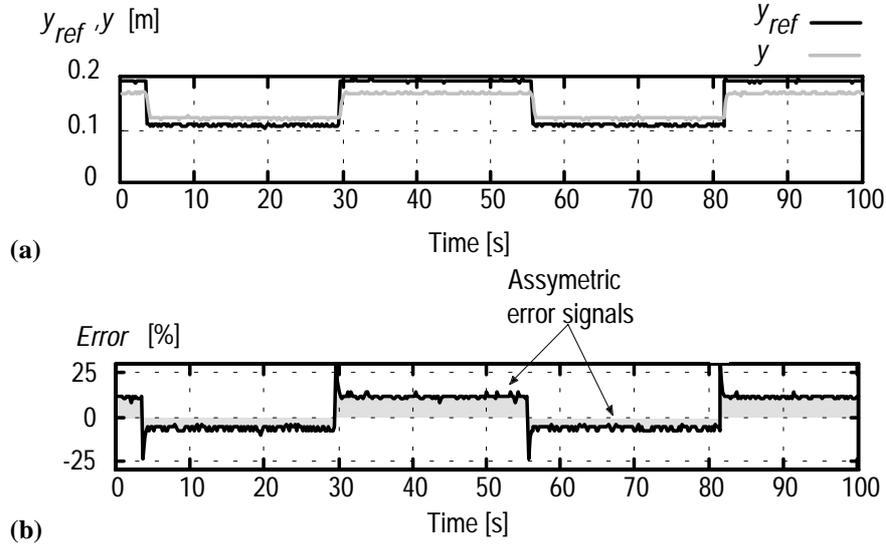

**(a)**

**(b)**

**Fig. 17: Experimental results obtained when the actuator operates with only the feedback-loop through the proportional controller. (a) Evolution of the reference signal ($y_{ref}$) and the piston position signal ($y$). (b) Asymmetric error evolution.**

The first results in Fig. 17 showed that the absence of compensator in the control-loop gives high tracking errors. In the second test, we added the compensation signal generated by the neuro-fuzzy inverse-model to the command signal of the proportional controller.

The results of the second test are displayed in figures 18a-b. They use the compensation feedforward-loop with the initial extracted neuro-fuzzy model but without the learning mechanism. These results show that the compensator eliminates the error signal in the superior part of the reference signal, but produces a higher error value in the inferior part. The compensation signal generated by the inverse relation was capable of distorting the proportional controller signal ($\omega_p$), thus increasing the error for the inferior part.

These results point out the necessity of more precise adjustment of model rules in the inferior operating region. Therefore, we introduce the neuro-fuzzy learning mechanism so the system acquires new signals in realtime and corrects the rules to tune the inverse model. The system uses the proportional controller signal to adjust, as described in Fig. 19, the rules of the inverse relation and then correct the compensation signal $\omega_{comp}$.

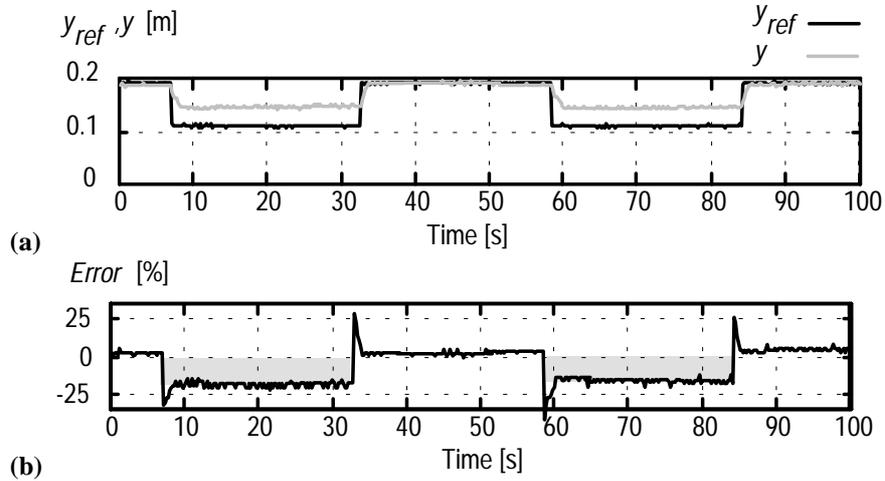

**Fig. 18: Experimental results when the feedforward-loop is added to the actuator system but without the neuro-fuzzy learning mechanism. (a) Evolution of the reference signal ($y_{ref}$) and the piston position signal ($y$). (b) Error signal evolution.**

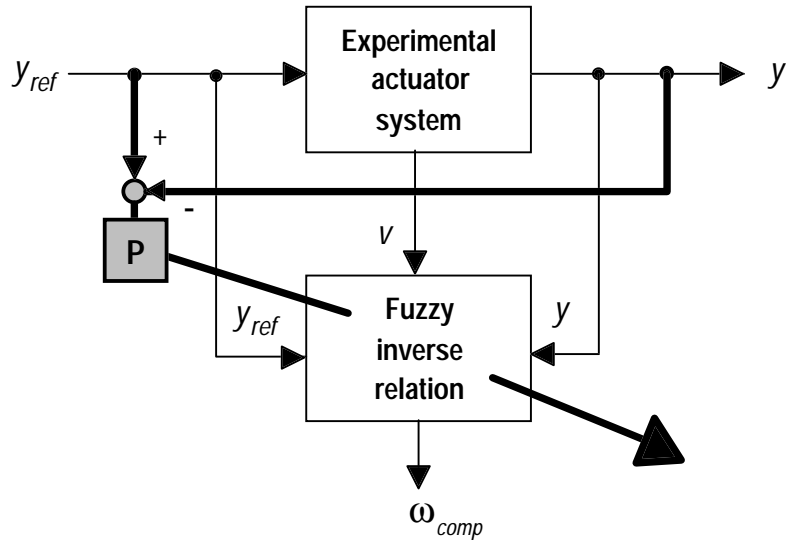

**Fig. 19: Diagram showing the use of the proportional controller signal to correct the inverse relation.**

The results in Fig. 20a illustrate the piston approximation to the reference signal as the rules are adjusted and the compensation signal is corrected. In this test, we used a low learning rate ($\alpha = 0.0005$) to better visualize the adjust of the compensation signal. As the learning mechanism begins to actuate, the system slowly increases the pump speed, as verified in Fig. 20b, to send fluid to the hydraulic circuit and so move the piston. The pump increases its speed until its magnitude becomes sufficient to remove the actuator out of the dead-zone, adjust the model rules, and then conduct the piston to the reference position reducing the error signal as shown in Fig. 20c.

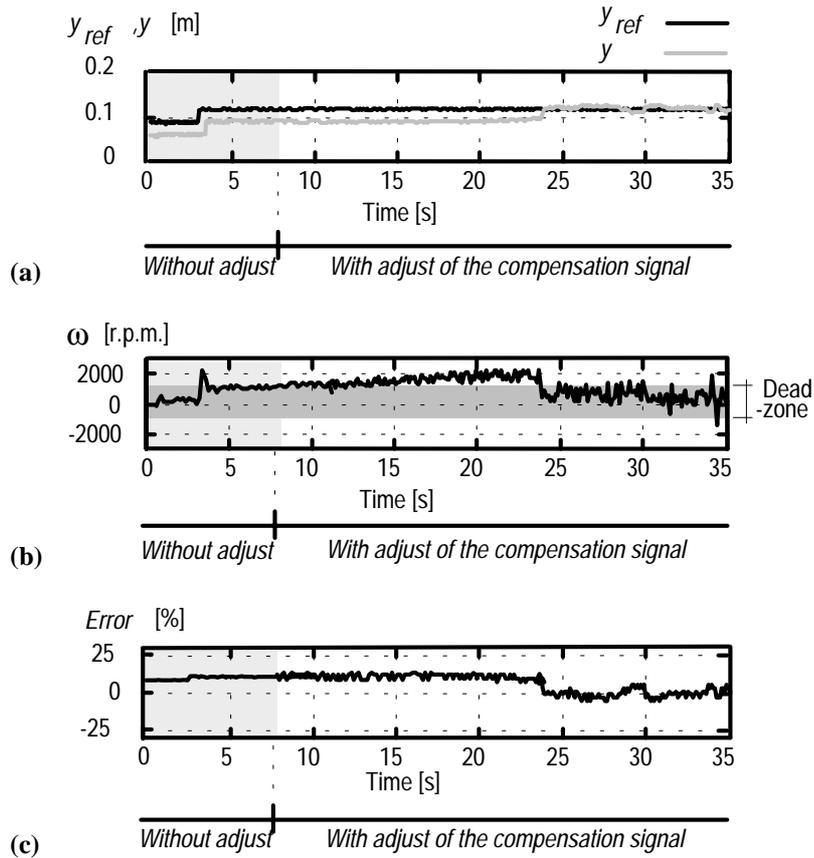

Fig. 20: Actuator's evolution for a step with the learning mechanism action to adjust the compensation signal. (a) Evolution of ($y_{ref}$) and piston position signal ($y$). (b) Hydraulic pump speed ($\omega$). (c) Error signal.

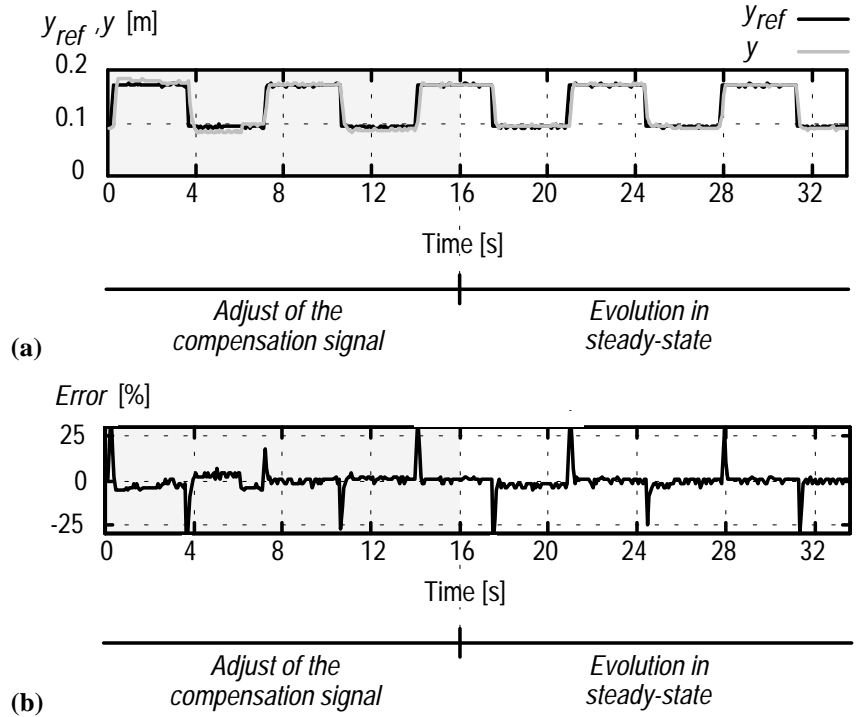

**Fig. 21: Actuator's evolution with the adjust, in realtime, of the model rules to correct the compensation signal. (a) Evolution of the reference signal ($y_{ref}$) and the piston position signal ($y$). (b) Error signal.**

In Fig. 21, we present the piston evolution when the feedforward-loop and the learning mechanism are inserted into the control system. This test uses a higher learning rate ($\alpha = 0.02$) for a fast transient but without overshoots. These results show the real-time tuning until about 16 seconds where the compensation signal gradually eliminates the error offset, approximating the piston to the reference signal.

# 8   Conclusion

The neuro-fuzzy methodology is used to demonstrate the incorporation of learning mechanisms into control of drive systems. We believe that emerging technologies as neuro-fuzzy systems have to be used together with usual conventional controllers to produce more "intelligent" and autonomous drive systems. All the knowledge accumulated about the classical controllers and emerging techniques as fuzzy systems, neural networks, or genetic algorithms should be utilized. So, it is becoming important to investigate control designs that permit a symbiotic effect between the old and new approaches. To the concretization and investigation of the anterior objectives, we presented a neuro-fuzzy modeling and learning approach to design a position controller for an electro-hydraulic actuator.

The results presented indicate the ability of the implemented neuro-fuzzy controller in performing learning and generalization properties to quite different movements than those presented during the training stage. The compensation of nonlinearities in the electro-hydraulic system deviating the feedback controller action to drive the piston position to its reference signal is also demonstrated.

# References

[1]   RayChaudhuri, T., Hamey, L.G.C., and Bell, R.D. (1996), From Conventional Control to Autonomous Intelligent Methods, *IEEE Control System Magazine*, Vol. 16, No. 5.
[2]   Zadeh, L. A. (1973), Outline of a New Approach to the Analysis of Complex Systems and Decision Processes, *IEEE Trans. on Systems, Man and Cybernetics*, Vol. SMC-3, No. 1, pp. 28-44.
[3]   Takagi, T. and Sugeno, M. (1985), Fuzzy Identification of Systems and Its Applications to Modeling and Control, *IEEE Tran. on Systems, Man, and Cybernetics*, Vol. SMC-15, No. 1, pp. 116-132.
[4]   Sugeno M. and Yasukawa, T. (1993), A Fuzzy-Logic-Based Approach to Qualitative Modeling, *IEEE Trans. on Fuzzy Systems*, Vol. 1, No. 1, pp. 7-31.
[5]   Myamoto, H., Kawato, M., Setoyama, T., and Suzuki, R. (1988), Feedback-Error-Learning Neural Network for Trajectory Control of a Robotic Manipulator, *Neural Networks* 1:251-265.
[6]   Wang, L. X. (1992), Back-Propagation of Fuzzy Systems as Nonlinear Dynamic System Identifiers, in *Proc. of IEEE Int. Conf. on Fuzzy Systems*, pp. 1409-1418, San Diego, CA, U.S.A.
[7]   Low, T-S., Lee, T-H., and Lim, H-K. (1993), A Methodology for Neural Network Training for Control of Drives with Nonlinearities, *IEEE Trans. on Industrial Electronics*, Vol. 39, No. 2, pp. 243-249.
[8]   Ohno, H., Suzuki, T., Aoki, K., Takahasi, A., and Sugimoto, G. (1994), Neural Network Control for Automatic Braking Control System, *Neural Networks*, Vol. 7, No. 8, pp. 1303-1312.
[9]   Miller III, W.T., Hewes, R.P., Glanz, F.H., and Kraft III, L.G. (1990), Real-Time Dynamic Control of an Industrial Manipulator Using a Neural-Network-Based Learning Controller", *IEEE Tran. on Robotics and Automation*, Vol. 6, No. 1, pp. 1-9.
[10]  Bondi, P., Casalino, G., and Gambardella, L. (1988), On The Iterative Learning Control Theory for Robotic Manipulators, *IEEE Journal of Robotics and Automation*, Vol. 4, No. 1, pp. 14-22, Feb.
[11]  Miller III, W.T. (1987), Sensor-Based Control of Robotic Manipulators Using a General Learning Algorithm, *IEEE Journal of Robotics and Automation*, Vol. RA-4, No. 2, pp. 157-165.

[12] Lewis, F.L. (1996), Neural Network Control of Robot Manipulators, *IEEE Expert*, pp. 65-75.
[13] Zadeh, L.A. (1965), Fuzzy Sets, *Information and Control*, Vol. 8, pp. 338-353.
[14] Sugeno, M. and Yasukawa, T. (1993), A Fuzzy-Logic-Based Approach to Qualitative Modeling, *IEEE Trans. on Fuzzy Systems*, Vol. 1, No. 1, pp. 7-31, Feb.
[15] Kawato, M., Uno, Y., Isobe, M., and Suzuki, R. (1988) Hierarchical Neural Network Model for Voluntary Movement with Application to Robotics, *IEEE Control System Magazine*, pp. 8-16, April.
[16] Lin, C.T. and Lee, C.S.G. (1991), Neural-Network-Based Fuzzy Logic Control and Decision System, *IEEE Trans. on Computers*, Vol. 40, No. 12, pp. 1320-1336.
[17] Wang, L.X. and Mendel, J.M. (1992), Fuzzy Basis Functions, Universal Approximation, and Orthogonal Least-Squares Learning, *IEEE Trans. on Neural Networks*, Vol. 3, No. 5, pp. 807-814.
[18] Costa Branco, P.J. and Dente, J.A. (1996), Inverse-Model Compensation Using Fuzzy Modeling and Fuzzy Learning Schemes. *In Intelligent Engineering Systems Through Artificial Neural Networks, Smart EngineeringSystems: Neural Networks, Fuzzy Logic and Evolutionary Programming,* Eds. C.H. Dagli, M. Akay, C.L. Philip Chen, B. Fernández, and J. Ghosh, Vol.6, pp. 237-242, ASME Press, New York, U.S.A.
[19] Wang, L.X. (1993), Training of Fuzzy-Logic Systems Using Neigborhood Clustering, *Proc. 2nd IEEE Conf. on Fuzzy Systems,* pp. 13-17, San Francisco, Califórnia, U.S.A., March.
[20] Wang, L.X. (1993), Stable Adaptive Fuzzy Control of Nonlinear Systems, *IEEE Trans. on Fuzzy Systems*, Vol. 1, No. 2, pp. 146-155.
[21] L.X. Wang, *Adaptive Fuzzy Systems and Control*, Englewood Cliffs, N.J.: Prentice-Hall, 1994.
[22] Holmblad, L.P. and Ostergaard, J.J. (1982), Control of Cement Kiln by Fuzzy Logic, M.M. Gupta and E. Sanchez, Eds. in *Approximate Reasoning in Decision Analysis*, Amsterdam: North-Holland. pp. 389-400.
[23] Mandani, E.H. (1977), Application of Fuzzy Logic to Approximate Reasoning Using Linguistic Variables, *IEEE Trans. on Computers*, Vol. C-26, pp. 1182-1191.
[24] Wang, L. X.. (1992), Back-Propagation of Fuzzy Systems as Nonlinear Dynamic System Identifiers, *Proc. IEEE Int. Conf. on Fuzzy Systems*, pp. 1409-1418, San Diego, CA.
[25] Pedrycz, W. (1984), An Identification of Fuzzy Relational Systems, *Fuzzy Sets and Systems*, Vol. 13, pp. 153-167.
[26] Czogala, E. and Pedrycz, W. (1981), On Identification in Fuzzy Systems and Its Applications in Control Problems", *Fuzzy Sets and Systems*, Vol. 6, No. 1, pp. 73-83.
[27] Costa Branco, P.J. and Dente, J.A. (1993), A New Algorithm for On-Line Relational Identification of Nonlinear Dynamic Systems", In *Proc. of Second IEEE Int. Conf. on Fuzzy Systems (IEEE-FUZZ'93)*, Vol. 2, pp. 1073-1079.
[28] Postlethwaite, B. (1991), Empirical Comparison of Methods of Fuzzy Relational Identification, *IEE Proc.-D*, Vol. 138, pp. 199-206.

[29] Higashi, M. and Klir, G.J. (1984), Identification of Fuzzy Relational Systems", *IEEE Trans. on Systems, Man, and Cybernetics*, Vol. SMC-14, No. 2, pp. 349-355.

[30] Xu, C.W. and Lu, Y.Z. (1987), Fuzzy Model Identification and Self-Learning for Dynamic Systems, *IEEE Trans. on Systems, Man, and Cybernetics*, Vol. SMC-17, No. 4.

[31] Costa Branco, P.J. and Dente, J.A. (1997), The Application of Fuzzy Logic in Automatic Modeling Electromechanical Systems, *Fuzzy Sets and Systems*, Vol. 95, No. 3, pp. 273-293.

[32] Costa Branco, P.J., Lori, N., and Dente, J.A. (1996), New Approaches on Structure Identification of Fuzzy Models: Case Study in an Electromechanical System, T. Furuhashi and Y. Uchikawa, Eds. in *Fuzzy Logic, Neural Networks, and Evolutionary Computation*, LNCS/Lecture Notes in Artificial Intelligence, Springer-Verlag, Berlin, pp. 104-143.

[33] Bezdek, J.C. and Pal, S.K., Eds. (1992), *Fuzzy Models For Pattern Recognition*, New York: IEEE Press.

[34] Tzafestas, S. and Papanikolopoulos, N.P. (1990), Incremental Fuzzy Expert PID Control, *IEEE Trans. on Industrial Electronics*, Vol. 37, No. 5, pp. 365-371.

[35] Ollero, A. and García-Cerezo, A.J. (1989), Direct Digital Control, Auto-Tuning and Supervision Using Fuzzy Logic, *Fuzzy Sets and Systems*, Vol. 30, pp. 135-153.

[36] Czogala, E. and Rawlik, T. (1989), Modeling of a Fuzzy Controller with Application to the Control of Biological Processes, *Fuzzy Sets and Systems*, Vol. 31, pp. 13-22.

[37] Jang, J-S. R. and Sun, C-T. (1995), Neuro-Fuzzy Modeling and Control, *Proceedings of the IEEE*, Vol. 83, pp. 378-405.

[38] Procyk, T.J. and Mamdani, E.H. (1979), A Linguistic Self Organising Process Controller, *Automatica*, Vol. 15, pp. 15-30.

[39] Shao, S. (1988), Fuzzy Self-Organising Controller and Its Application for Dynamic Processes, *Fuzzy Sets and Systems*, Vol. 26, pp. 151-164.